\documentclass{article}

\usepackage{microtype}
\usepackage{graphicx}
\usepackage{subcaption}
\usepackage{booktabs} % for professional tables
\usepackage{cleveref}

\usepackage[accepted]{icml2019}

\usepackage{url}

\usepackage{breakurl}
\usepackage[font=small]{caption}
\usepackage{wrapfig}
\usepackage{textcomp}
\usepackage{blindtext}

\usepackage{multirow}
\usepackage{dsfont}
\usepackage{multicol}
\usepackage{nicefrac}       % compact symbols for 1/2, etc.
\usepackage{amsmath}
\usepackage{amsthm}
\usepackage{amssymb}
\usepackage{bm}
\usepackage{mathrsfs}
\usepackage{xcolor}
\usepackage{colortbl}
\usepackage[hang,flushmargin]{footmisc}

\usepackage{color}
\usepackage{soul}

\usepackage{pgfplots}
\pgfplotsset{
  compat=newest,
  plot coordinates/math parser=false,
  tick label style={font=\footnotesize, /pgf/number format/fixed},
  label style={font=\small},
  legend style={font=\small},
  every axis/.append style={
    tick align=outside,
    clip mode=individual,
    scaled ticks=false,
    thick,
    tick style={semithick, black}
  }
}

\pgfkeys{/pgf/number format/.cd, set thousands separator={\,}}

\usetikzlibrary{external}
\newlength\figurewidth
\newlength\figureheight

\newlength\sbsfigurewidth
\newlength\sbsfigureheight

\usepackage{xspace}

\newcommand{\mc}[1]{\mathcal{#1}}
\newcommand{\data}{\mc{D}}
\newcommand{\model}{\mc{M}}
\newcommand{\given}{\mid}

\newcommand{\intd}[1]{\,\mathrm{d}{#1}}

\newcommand{\acro}[1]{\textsc{\MakeLowercase{#1}}}

\DeclareMathOperator*{\argmax}{arg\,max}

\icmltitlerunning{Automated Model Selection with Bayesian Quadrature}

\begin{document}

\twocolumn[
\icmltitle{Automated Model Selection with Bayesian Quadrature}


\icmlsetsymbol{equal}{*}

\begin{icmlauthorlist}
\icmlauthor{Henry Chai}{wustl}
\icmlauthor{Jean-Fran\c{c}ois Ton}{oxstat}
\icmlauthor{Roman Garnett}{wustl}
\icmlauthor{Michael A. Osborne}{oxeng}
\end{icmlauthorlist}

\icmlaffiliation{wustl}{Department of Computer Science and Engineering, Washington University in St. Louis, St. Louis, MO, USA}
\icmlaffiliation{oxstat}{Department of Statistics, University of Oxford, Oxford, United Kingdom}
\icmlaffiliation{oxeng}{Department of Engineering Science, University of Oxford, Oxford, United Kingdom}

\icmlcorrespondingauthor{Henry Chai}{hchai@wustl.edu}
\icmlkeywords{Bayesian Quadrature, Model Selection, Active Learning, Gaussian Processes}

\vskip 0.3in
]



\printAffiliationsAndNotice{}  % leave blank if no need to mention equal contribution

\begin{abstract}
	% !TEX root = ../main.tex

We present a novel technique for tailoring Bayesian quadrature (\acro{BQ}) to model 
selection. The state-of-the-art for comparing the evidence of multiple models relies 
on Monte Carlo methods, which converge slowly and are unreliable for computationally 
expensive models. Previous research has shown that \acro{BQ} offers sample efficiency 
superior to Monte Carlo in computing the evidence of an individual model. However, 
applying \acro{BQ} directly to model comparison may waste computation producing an 
overly-accurate estimate for the evidence of a clearly poor model. We propose an 
\emph{automated and efficient} algorithm for computing the most-relevant quantity 
for model selection: the posterior probability of a model. Our technique maximizes 
the mutual information between this quantity and observations of the models' likelihoods, 
yielding efficient acquisition of samples across disparate model spaces when likelihood 
observations are limited. Our method produces more-accurate model posterior estimates 
using fewer model likelihood evaluations than standard Bayesian quadrature and Monte 
Carlo estimators, as we demonstrate on synthetic and real-world examples.
\end{abstract}

% !TEX root = ../main.tex

\section{Introduction}
\label{intro}
Model selection is a fundamental problem that arises in the course of scientific inquiry: 
which of several candidate models best explains an observed dataset $\data$? The Bayesian 
approach to model selection involves computing \emph{posterior model probabilities,} the 
probability that each model generated the observations. This approach requires computing 
\emph{model evidences,} which can be expressed as integrals of the form $Z = \int 
f(\data\mid\theta)\,\pi(\theta)\intd{\theta},$ where $\theta$ is a vector of model 
parameters, $f(\data \mid \theta)$ is a likelihood, and $\pi(\theta)$ is a prior. 

Unfortunately, for many real-world model selection tasks, these integrals are 
computationally intractable and must be estimated numerically. Numerous commonly-used 
techniques to estimate such integrals rely on \emph{Monte Carlo} estimators 
\citep{metropolis53, hastings70}. These methods converge slowly in terms of the number of 
required integrand samples as they do not incorporate knowledge about sample locations.
This makes such methods ill-suited for settings where the integrand is expensive to 
evaluate. 

One alternative is \emph{Bayesian quadrature} (\acro{BQ}) 
\citep{larkin72, diaconis88, hagan91, rasmussen03}, which relies on a probabilistic 
belief on the integrand that can be conditioned on observations to derive a posterior 
belief about the value of the integral. The theoretical properties of kernel quadrature 
methods (including \acro{BQ}) have been studied at length: these methods can achieve 
faster convergence rates than Monte Carlo estimators \citep{briol15, bach17, karvonen18}, 
even when the underlying model is misspecified \citep{kanagawa16, kanagawa17}, a 
commonly-cited pitfall of kernel-based methods.

There has been significant work investigating how traditional, Monte Carlo based methods 
can be adapted to efficiently estimate posterior model probabilities for model selection 
\citep{neal01, green95, chib95, meng96, godsill01, skilling04}. However, no analogous work 
has appeared to adapt \acro{BQ} for this important task. That is our goal in this work. 

We propose a principled adaptations of \acro{BQ} designed to automate model selection. 
Specifically, we define a novel acquisition functions for active selection of locations to 
observe model likelihoods. This acquisition functions corresponds to the mutual 
information between observations of the model likelihood and a quantity specifically 
relevant to the task of model selection: the posterior model probabilities. This allows 
our method to automatically select informative sample locations across multiple model 
parameter spaces unlike previous active \acro{BQ} approaches to model selection 
\citep{osborne12, gunter14, chai18}, which focused on accurately estimating individual 
model evidences. We illustrate the shortcomings of such approaches using toy motivating 
examples. Experiments conducted on real-world and synthetic data demonstrate that our 
proposed method can outperform previously proposed \acro{BQ} techniques and specialized 
Monte Carlo methods in terms of efficiently reaching accurate estimates of posterior 
model probabilities. 
% !TEX root = ../main.tex

\section{Related Work}
\label{rel}
Much work has been devoted to developing Monte Carlo methods specifically designed for 
model selection. Broadly speaking, these methods can be broken down into two groups: 
within-model approaches, such as annealed importance sampling (\acro{AIS}) \citep{neal01}, 
nested sampling \citep{skilling04}, and bridge sampling \citep{bennett76,meng96}, and 
trans-dimensional approaches such as \citet{green95}'s reversible jump \acro{MCMC} and 
\citet{godsill01}'s composite model space framework, a generalization of the product-space 
approach proposed by \citet{chib95}. Within-model approaches estimate each model's model 
evidence separately, whereas trans-dimensional approaches directly estimate posterior 
model probabilities.

In our experiments we compare our method against one prominent Monte Carlo method from 
each category: bridge sampling (within-model) and reversible jump \acro{MCMC} 
(trans-dimensional). All commonly used Monte Carlo methods for model selection have pros 
and cons, and their performance on specific tasks can be greatly affected by open-ended 
modeling choices such as the choice of intermediate densities for \acro{AIS} or the 
choice of pseudo-priors for the composite model space framework of \citet{godsill01}. It 
is beyond the scope of this work to comprehensively analyze all widely used Monte Carlo 
model selection methods; however, we do believe the benchmarks we choose to be reasonable 
and competitive. In particular, the design choices associated with bridge sampling are 
easily justified and give rise to greater transparency in our experimental design as 
opposed to many possible alternatives. 

Among the commonly used trans-dimensional methods, the original product space approach of 
\citet{chib95} made use of a Gibbs sampler, which requires conjugate conditional 
likelihoods that do not exist in many model selection settings. \citet{godsill01} showed 
that replacing the Gibbs sampler in their composite space model with a 
Metropolis--Hastings proposal mechanism gives rise to \citet{green95}'s reversible jump 
\acro{MCMC}. \citet{godsill01} also claimed that the use of such a Metropolis--Hastings 
proposal mechanism is preferable to the use of a Gibbs sampler in nested model settings, 
that is, settings where there is an overlap in model parameter spaces. As our real-world 
experimental setting is a model selection task between nested models, the choice to 
compare against reversible jump \acro{MCMC} is well justified.  

Previous work has been done on adapting \acro{BQ} to situations where the integrand of an 
intractable integral is known to be nonnegative \emph{a priori} 
\citep{osborne12, gunter14, chai18}. Such integrals occur frequently in machine learning 
tasks, including model selection: the model evidence is an integral of probability 
distributions, which are nonnegative everywhere. These methods make use of warped 
\acro{GP}s \citep{snelson04} to weakly enforce the nonnegativity constraint. They have 
been shown to outperform \acro{BQ} algorithms that are agnostic to \emph{a priori} 
information on a variety of model selection tasks. However, we will show that our method 
can lead to even greater improvements. Furthermore, our methodology is compatible with 
the use of warped \acro{GP}s; in fact, our proposed method can be seen as an instantiation 
of the framework laid out by \citet{chai18} with a novel acquisition function. 

As a final note, our focus in this manuscript is on the traditional Bayesian approach to 
model selection, which involves the computation of model posteriors. We acknowledge the 
existence of several alternative approaches to Bayesian model selection 
\citep{bernardo94, watanabe10, vehtari12}. An extension of our proposed method to these 
alternatives is a potential line of future inquiry. 
% !TEX root = ../main.tex

\section{Background}
\label{back}

\subsection{Bayesian Model Selection}
\label{ms}

For the purposes of this work, a model is defined as a parametric family of probability 
distributions that can be used to explain some observed dataset, $\data$. Given a finite 
set of models candidate models $\{\model_1,\ldots,\model_k\}$, the Bayesian approach to 
inference in this setting is to reason about the conditional or posterior distribution 
over models via Bayes theorem: 
\begin{align}
    \Pr(\model_i\given\data)&=\frac{\Pr(\data\given\model_i)\Pr(\model_i)}{\Pr(\data)} \nonumber \\ 
    						&=\frac{\Pr(\data\given\model_i)\Pr(\model_i)}{\sum_{j=1}^k \Pr(\data\given\model_j)\Pr(\model_j)}
    \label{eq1}
\end{align}
where $\Pr(\model_i\given\data)$ is known as the posterior probability of model 
$\model_i$, $\Pr(\data\mid\model_i)$ is the model evidence of model $\model_i$ and
$\Pr(\model_i)$ is the prior probability of model $\model_i$. The computation of model 
evidences requires integrating out the model parameters that control the likelihood of a 
given model generating the observed data:
\begin{equation}
    \Pr(\data\given\model_i)=\int\Pr(\data\given\model_i,\theta_i)\Pr(\theta_i\given\model_i)\intd{\theta_i}
    \label{eq2}
\end{equation}
where $\theta_i$ is the vector of model parameters corresponding to model $\model_i$, 
$\Pr(\data\given\model_i,\theta_i)$ is the likelihood of $\data$ under $\model_i$ 
parameterized by $\theta_i$, and $\Pr(\theta_i\given\model_i)$ is the prior probability 
of the model parameters $\theta_i$. 

Given posterior model probabilities, one common practice is to find 
$\model^*=\argmax_{\model}\Pr(\model_i\given\data)$ and then treat $\model^*$ as the true, 
data-generating model for the purpose of future inference tasks. We will refer to this 
variant of the model selection task as \emph{model choice}. An alternative to model choice 
is \emph{model averaging}: instead of simply using the single, most-likely candidate 
model, model averaging takes a fully-Bayesian viewpoint by using the posterior model 
probabilities to marginalize out the choice of model for subsequent inference tasks. 

\subsection{Mutual Information}
\label{mi}

The \emph{mutual information} between two random variables is a measure of how much 
information observing the value of one provides about the other. Formally, given two 
random variables $X$ and $Y$, the mutual information of $X$ and $Y$ is
\begin{equation}
  I(X;Y) = H(X)-H(X\mid Y),
  \label{eq3}
\end{equation}
where $H(X)$ is the \emph{entropy} of $X$, and $H(X \mid Y)$ is the 
\emph{conditional entropy} of $X$ given $Y$. If $X$ is discrete with \acro{PMF} $p$ and 
domain $\mathcal{X}$, then the entropy is defined to be
\begin{equation}
  H(X)=-\sum_{x\in\mathcal{X}}p(x)\log p(x)
  \label{eq4}
\end{equation}
and if $X$ is continuous with \acro{PDF} $p$ and domain $\mathcal{X}$, then we instead 
use the \emph{differential entropy:}
\begin{equation}
  H(X)=-\int_{\mathcal{X}}p(x)\log p(x)\intd{x}.
  \label{eq5}
\end{equation}
The conditional entropy $H(X \mid Y)$ is defined to be the expected (differential) 
entropy of the posterior distribution $p(X \mid Y)$, where the expectation is taken with 
respect to $Y$. Therefore the mutual information can be interpreted as the expected 
information gained about $X$ (that is, the expected reduction in entropy) when measuring 
$Y$.

\subsection{Bayesian Quadrature}
\label{bq}

Given some intractable integral $Z=\int f(\theta)\,\pi(\theta)\intd\theta$, Bayesian 
quadrature \acro{BQ} places a Gaussian process (\acro{GP}) prior belief on the function 
$f(\theta)$ (or occasionally on the product $f(\theta)\,\pi(\theta)$ directly). A 
\acro{GP} specifies a probability distribution over functions, where the joint 
distribution of the function's value at finitely many locations is multivariate normal. 
Much like a multivariate Gaussian distribution is fully specified by its first two 
moments, a \acro{GP} is fully specified by its first two moments: a mean function $\mu(x)$ 
and a covariance function $\Sigma(x,x')$. Given a set of observations at locations 
$x_D=\{x_1,\ldots,x_n\}$ with corresponding function values $f(x_D)$, a GP prior can be
conditioned on these observations to arrive at a posterior GP with mean 
$m_D(x)=m(x)+K(x,x_D)K(x_D,x_D)^{-1}(f(x_D)-m(x_D))$ and covariance 
$K_D(x,x')=K(x,x')-K(x,x_D)K(x_D,x_D)^{-1}K(x_D,x')$. For a comprehensive overview of 
\acro{GP}s, see \citep{rasmussen06}.

\acro{BQ} makes use of the fact that \acro{GP}s are closed under linear functionals 
\citep{rasmussen03}, meaning that a \acro{GP} belief on $f$ induces a Gaussian belief on 
$L[f]$, where $L$ is any linear functional. As integration against a probability 
distribution is such a functional, then if $p(f) = \mathcal{GP}(f; \mu,\Sigma)$, we have 
$p(Z) = N(Z; m,K)$, where
\begin{align}
  m&=\int\mu(x)\,\pi(x)\intd{x};\\
  K&=\iint\Sigma(x,x')\,\pi(x)\,\pi(x')\intd{x}\intd{x'}.
  \label{eq6}
\end{align}
A design choice that must be addressed when using \acro{BQ} is where to observe the 
integrand $f$. One natural approach is to select sample locations so as to minimize one's
uncertainty about $Z$ which is equivalent to minimizing the entropy of $Z$. Because the 
posterior variance of a \acro{GP} does not depend on the observed function values 
(see above), if a \acro{GP} prior is placed directly on $f$, then an optimal sampling 
design (w.r.t. this objective) can be specified in advance \citep{minka00}.

However, it is often appropriate to specify a \acro{GP} prior not on $f$ but on an affine 
transformation of $f$ in order to incorporate some \emph{a priori} information. Doing so 
introduces a dependency between the posterior variance and observed function values. 
Thus, the optimal sampling sequence cannot be precomputed. One option in this setting is 
to sequentially selected in order to minimize the expected entropy of $Z$ as proposed by 
\citet{osborne12}. As an alternative, \citet{gunter14} propose an active sampling 
mechanism that iteratively minimizes the entropy of the integrand instead of the value of 
the integral as doing so is more computationally efficient and numerically stable.
% !TEX root = ../main.tex

\section{Motivation}
\label{mo}

\begin{figure}[!h]
    \includegraphics{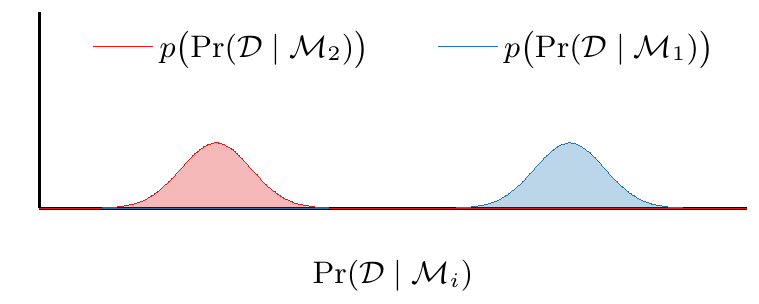}
    \caption{A toy example of the posteriors for the evidences of two models; observe that 
    both posterior beliefs are Gaussian, a direct result of the closure of \acro{GP}s 
    under linear functionals (see \eqref{eq6})} 
    \label{fig0}
\end{figure}

Consider the task of selecting between two models, $\model_1$ and $\model_2$, given data 
$\data$. Suppose that \acro{BQ} is used to estimate the model evidences for both models. 
After some number of iterations of \acro{BQ}, the posterior beliefs (implicitly 
conditioned on \acro{BQ} evaluations) on the model evidences are plotted on the same 
axis; as an example, see \Cref{fig0}.

\Cref{fig0} depicts a situation where there is high uncertainty about both model 
evidences; however, the uncertainty in the posterior model probabilities is low. In 
particular, for this toy example, $\Pr(\model_1\given\data)$ is almost certainly close to 
one, while $\Pr(\model_2\given\data)$ is almost certainly close to zero. This example 
illustrates the fact that it is not necessary to have low-entropy estimates of model 
evidences to have low-entropy estimates of posterior model probabilities and thus, 
methods that aim to achieve accurate estimates of model evidences may be inefficiently 
sampling observations when the goal to is to achieve accurate estimates of the posterior 
model probabilities.
% !TEX root = ../main.tex

\section{Methods}
\label{meth}
We propose an adaptation of the standard \acro{BQ} algorithm to the task of model 
selection. The principal novelty of is in selecting where to observe the likelihood 
functions of the models involved. Rather than selecting locations with the goal of 
achieving accurate estimates of the model evidences, as previous work has considered at 
length \citep{osborne12, gunter14}, our method seeks to achieve accurate estimates of 
posterior model probabilities, a more essential quantity to model selection. Our method 
makes use of the mutual information between model likelihood observations and posterior 
model probabilities, allowing them to choose informative sample locations across multiple 
model parameter spaces simultaneously. 

In the description that follows, we suppose that we have observed some dataset $\data$ 
and have two candidate models, $\model_1$ and $\model_2$, to explain $\data$. Let 
$\ell_i(\theta_i)=p(D \mid\theta_i,\model_i)$ be the likelihood for $\model_i$. We assume 
that these likelihoods have mutually independent Gaussian process priors: 
$p(\ell_i)=\mathcal{GP}(\ell_i; \mu_i,\Sigma_i)$. Now the model evidence of $\model_i$:
\begin{equation}
    a_i=\int \ell_i(\theta_i)\,\pi_i(\theta_i)\intd\theta_i 
    \label{eq7}
\end{equation}
is normally distributed as $p(a_i) = \mc{N}(a_i; m_i, K_i)$, where $m_i$ and $K_i$ are 
given by the standard \acro{BQ} identities \eqref{eq6}. 

Formally, we consider the mutual information between $z_1$ and $\ell_j(\theta_j)$. We 
first present the mutual information between $z_1$ and $\ell_1(\theta_1)$ (the result for 
the mutual information between $z_1$ and $\ell_2(\theta_2)$ is very similar and shown at 
the end of this section),
\begin{equation}
    I \bigl(\ell_1(\theta_1);z_1\bigr)=H\bigl(\ell_1(\theta_1)\bigr)-H\bigl(\ell_1(\theta_1)\given z_1\bigr).
    \label{eq8}
\end{equation}
Here $\ell_1(\theta_1)$ is just a univariate Gaussian and its entropy is 
$H(\ell_1(\theta_1))=\nicefrac{1}{2}\log 2\pi e\,\Sigma_1(\theta_1,\theta_1)$. 
Interestingly, the conditional random variable $\ell_1(\theta_1)\given z_1$ is also a 
univariate Gaussian. To see this, consider the joint density between $\ell_1(\theta_1)$, 
$a_1$, $a_2$, and $b_1=(z_1-1)a_1+z_1a_2$. As $a_1$ and $a_2$ are both independent and 
Gaussian, $b_1$ is jointly Gaussian with $a_1$ and $a_2$: 
$p(b_1) = N(b_1;\beta_1,s_1^2)$, where $\beta_1=(z_1-1)m_1+z_1m_2$ and 
$s_1^2=(z_1-1)^2K_1+z_1^2K_2$. Additionally, because $a_1$ is a linear functional of 
$\ell_1$ and there is a \acro{GP} belief on $\ell_1$, $\ell_1(\theta_1)$ and $a_1$ are 
also jointly Gaussian. More precisely: 
\begin{multline}
    p\left\{\begin{bmatrix}l_1(\theta_1) \\ a_1 \\ a_2 \\ b_1 \end{bmatrix}\bigm|\theta_1\right\}=N\left(\begin{bmatrix}\mu_1(\theta_1) \\ m_1 \\ m_2 \\ \beta_1 \end{bmatrix}, \right. \nonumber \\
    \left. \begin{bmatrix}\Sigma_1(\theta_1,\theta_1) & L_1(\theta_1) & 0 & (z_1-1)L_1(\theta_1) \\ L_1(\theta_1) & K_1 & 0 & (z_1-1)K_1 \\ 0 & 0 & K_2 & z_1K_2 \\ (z_1-1)L_1(\theta_1) & (z_1-1)K_1 & z_1K_2 & s_1^2 \end{bmatrix}\right)
    \label{eq9}
\end{multline}
where 
\begin{equation}
    L_1(\theta_1)=\int\Sigma_1(\theta_1,\theta_1')\,\pi_1(\theta_1')\intd{ \theta_1'}.
    \label{eq10}
\end{equation}
We note that observing $z_1$ is equivalent to observing that $b_1=0$; in essence, this 
follows because an observation of $z_1$ collapses the joint Gaussian distribution between 
$a_1$ and $a_2$ down to a line, where each point with support under the conditional 
belief satisfies the invariant that $(z_1-1)a_1+z_1a_2=0$. Thus, we can conclude that 
$\ell_1(\theta_1)\given z_1$ and $\ell_1(\theta_1)\given b_1=0$ have the same 
distribution. Using the fact that Gaussians are closed under conditioning and 
marginalization, it follows that
\begin{multline}
    p\bigl(\ell_1(\theta_1)\given\theta_1,b_1=0\bigr)=\\
    N\biggl(\mu_1(\theta_1)-\beta_1\frac{(z_1-1)L_1(\theta_1)}{s_1^2},\\
    \!\!\Sigma_1(\theta_1,\theta_1)-\frac{(z_1-1)^2L_1(\theta_1)^2}{s_1^2}\biggr).
    \label{eq11}
\end{multline}
Therefore, $\ell_1(\theta_1)\given z_1$ is a Gaussian random variable. The entropy of the 
posterior distribution $p\bigl(\ell_1(\theta_1) \given z_i, \theta_1)$ is
\begin{align}
    &\tfrac{1}{2}\log\Sigma_1(\theta_1,\theta_1) \nonumber \\ {}-&\tfrac{1}{2}\log\left(\Sigma_1(\theta_1,\theta_1)- \frac{(z_1-1)^2L_1(\theta_1)^2}{s_1^2}\right).
    \label{eq12}
\end{align}
Following a similar train of reasoning the entropy of the posterior distribution 
$p\bigl(\ell_2(\theta_2) \given z_i, \theta_2)$ is
\begin{align}
    &\tfrac{1}{2}\log\Sigma_2(\theta_2,\theta_2) \nonumber \\ {}-&\tfrac{1}{2}\log\left(\Sigma_2(\theta_2,\theta_2)- \frac{z_1^2L_2(\theta_2)^2}{s_1^2}\right).  
    \label{eq13}
\end{align} 
The mutual information is now the expectation of these quantities over $z_1$:
\begin{align}
    I(\ell_1&(\theta_1);z_1)=\tfrac{1}{2}\log\Sigma_1(\theta_1,\theta_1) \nonumber \\ &-\tfrac{1}{2}\int\log\left(\Sigma_1(\theta_1,\theta_1)- \frac{(z_1-1)^2L_1(\theta_1)^2}{s_1^2}\right)\,p(z_1)\intd{z_1} \label{eq14} \\
    I(\ell_2&(\theta_2);z_1)=\tfrac{1}{2}\log\Sigma_2(\theta_2,\theta_2)\nonumber \\ &-\tfrac{1}{2}\int\log\left(\Sigma_2(\theta_2,\theta_2)- \frac{z_1^2L_2(\theta_2)^2}{s_1^2}\right)\,p(z_1)\intd{z_1}.
    \label{eq15}
\end{align} 
Now to design our next observation, we choose to evaluate the likelihood at the point 
maximizing the expected information gain about $z_1$, where we maximize over the parameter 
spaces of \emph{both models.} That is, defining 
\begin{equation}
    \theta_i^* = \argmax_{\theta_i} I\bigl(z_1;\ell_i(\theta_i)\bigr),
    \label{eq16}
\end{equation}
for $i \in \{1,2\}$, we choose the next evaluation at $\theta_1^*$ iff 
\begin{equation}
    I\bigl(z_1;\ell_1(\theta_1^*)\bigr) > I\bigl(z_1;\ell_2(\theta_2^*)\bigr).    
    \label{eq17}
\end{equation}
Otherwise, we choose $\theta_2^*$.

Unfortunately, the integrals in \eqref{eq14} and \eqref{eq15} are intractable. Luckily, 
these can be accurately and efficiently estimated using standard numerical techniques. 
Also note that the estimation of these integrals is only performed as a means of selecting 
likelihood observation locations; the effect of inaccuracies in this estimate on the 
estimated model evidences will be negligible. 

\begin{figure*}[!t]
    \captionsetup[subfigure]{labelformat=empty}
    \centering
    \begin{subfigure}{0.49\textwidth}
      	\includegraphics{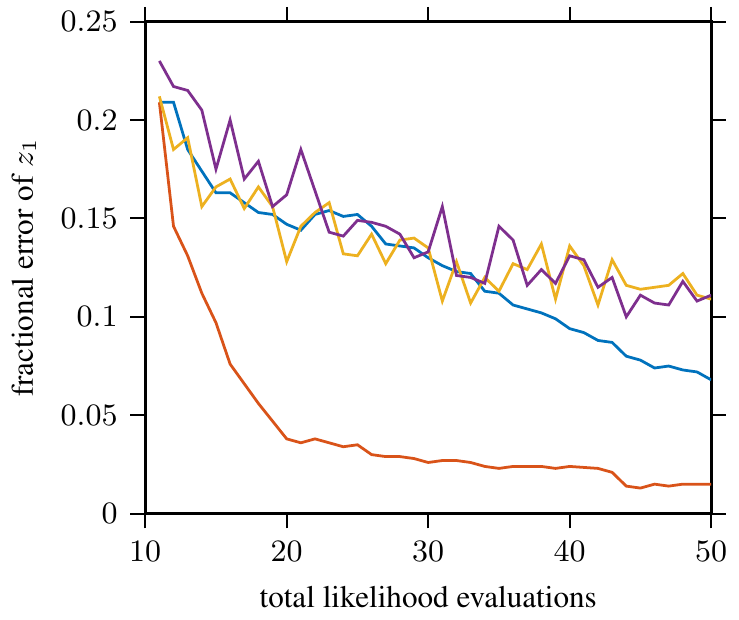}
        \caption{(a) $d=1$}
        \label{fig2a}
    \end{subfigure}
    \begin{subfigure}{0.49\textwidth}
      	\includegraphics{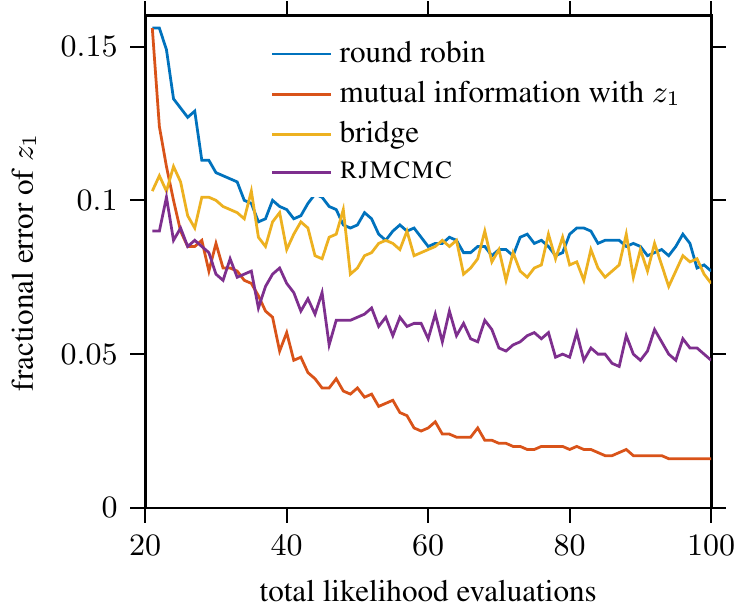}
        \caption{(b) $d=2$}
        \label{fig2b}
    \end{subfigure} \\
    \begin{subfigure}{0.49\textwidth}
      	\includegraphics{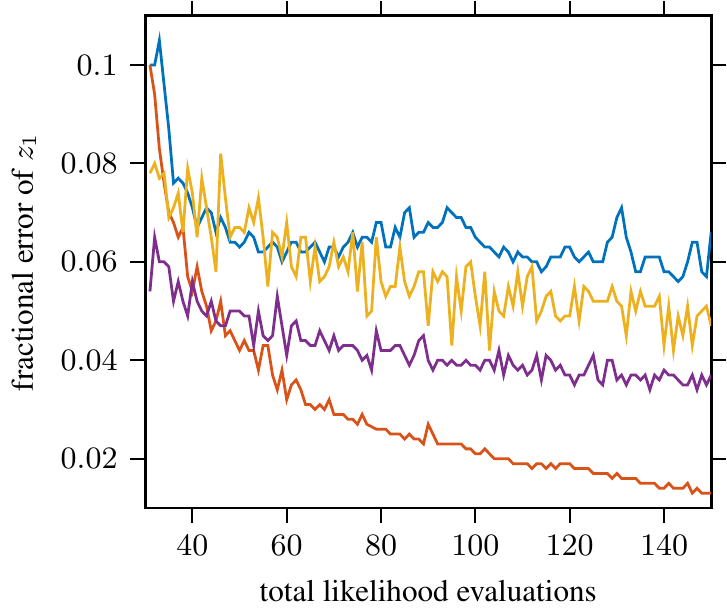}
        \caption{(c) $d=3$}
        \label{fig2c}
    \end{subfigure}
    \begin{subfigure}{0.49\textwidth}
      	\includegraphics{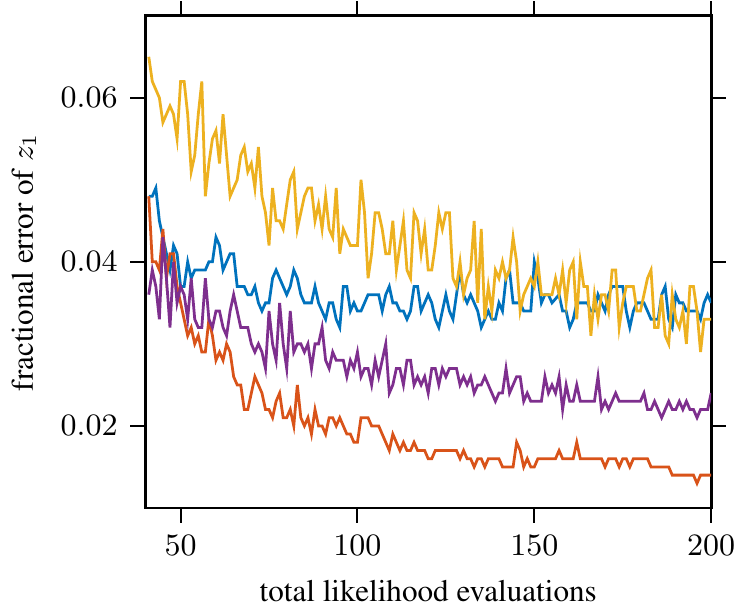}
        \caption{(d) $d=4$}
        \label{fig2d}
    \end{subfigure} \\
    \begin{subfigure}{0.49\textwidth}
      	\includegraphics{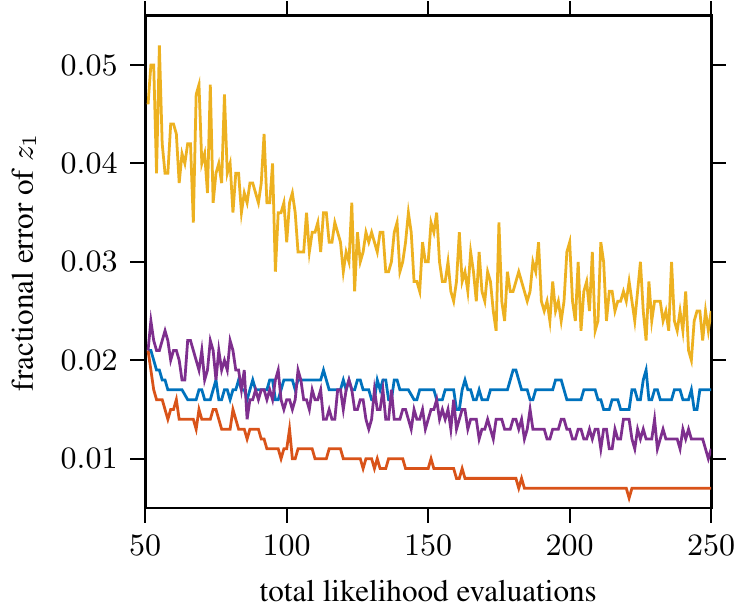}
        \caption{(e) $d=5$}
        \label{fig2e}
    \end{subfigure}
    \caption{Fractional errors of $z_1$ for all tested methods as a function of the total number of model likelihood observations}
    \label{fig2}
\end{figure*}

% note that the quantities $I(\ell_1(\theta_1);z_1)$ and $I(\ell_2(\theta_2);z_1)$ are both differentiable w.r.t.\ $\theta_1$ and $\theta_2$, respectively (for most common choices of covariance functions $\Sigma_1$ and $\Sigma_2$) so these quantities can be optimized using gradient-based methods.
% !TEX root = ../main.tex

\section{Experiments}
\label{exps}

We perform experiments on synthetic and real-world data in which we compare our proposed 
method against round-robin \acro{BQ}, where likelihood evaluations are evenly distributed 
between all model parameter spaces, and two Monte Carlo based benchmarks: bridge sampling 
\citep{meng96} and reversible jump \acro{MCMC} \citep{green95}. Our implementation of 
bridge sampling follows the one described by \citet{gronau17}: specifically, we use the 
optimal bridge function defined by \citet{meng96} and a Gaussian proposal distribution 
with moments fit to samples from the true posterior distribution (as suggested by 
\citet{overstall10}). Our choice of diffeomorphism for reversible jump \acro{MCMC} varies 
by experimental setting and is described in the relevant sections below. For all \acro{BQ} 
methods, constant-mean \acro{GP} priors with Mat\'{e}rn covariance functions 
($\nu=\nicefrac{3}{2}$) were placed on the log of the model likelihoods and all \acro{GP} 
hyperparameters were fit in accordance with the framework defined by \citet{chai18}. Our 
implementation of round-robin \acro{BQ} uses uncertainty sampling to select locations to 
observe log-likelihoods, as proposed by \citet{gunter14}. 

\subsection{Synthetic Experiments}
\label{synth}
For our synthetic experiments, we consider a model selection task between two zero-mean 
\acro{GP} models: one chosen to have a squared exponential covariance and one chosen to 
have a Mat\'{e}rn covariance with $\nu=\nicefrac{5}{2}$. The observed dataset $\data$ 
consists of $5d$ observations from a $d$-dimensional, zero-mean \acro{GP} with a squared 
exponential covariance. Each model is parameterized by the $d$ length-scales of their 
respective covariance functions (for the sake of simplicity, all other \acro{GP} 
hyperparameters were set to be the same as the true, data-generating \acro{GP}'s). 
In this setting, prior knowledge of the experimental setting suggests that the two 
likelihood functions are similar. Thus an appropriate choice of diffeomorphism is the 
identity function and the corresponding Jacobian is always 1. The intractable integrals
associated with our proposed method are estimated using 10000 simple Monte Carlo 
(\acro{SMC}) samples i.e. samples drawn from the probability distribution being 
integrated against. 

We allot a budget of $50d$ total likelihood evaluations and initialize each \acro{BQ} 
based method with $5d$ randomly sampled likelihood observations from both model parameter 
spaces. We run experiments with $d$ ranging from 1 to 5 and for each value of $d$, we 
consider 100 different, randomly sampled observed datasets. All methods are evaluated on 
the fractional error of their $z_1$ estimates with ground truth values being determined 
by exhaustive \acro{SMC}.

\begin{figure}[!h]
    \includegraphics{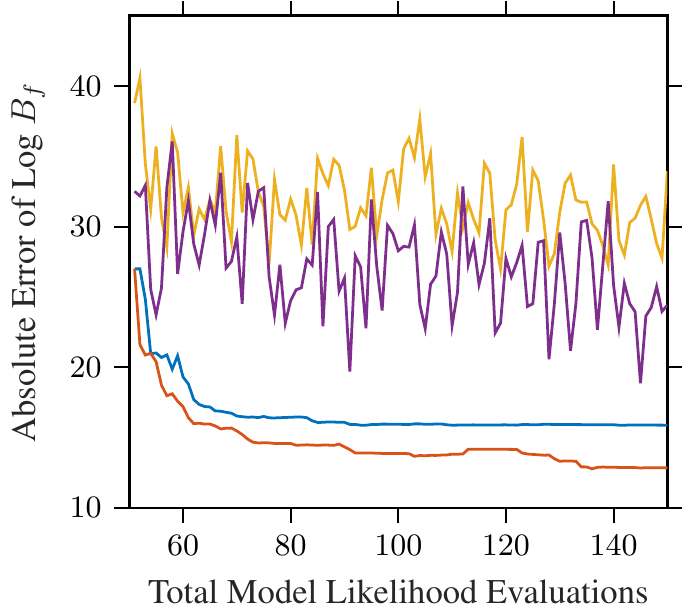}
    \caption{Absolute errors of log odds for all tested methods as a function of the 
    total number of model likelihood observations. This figure shares a legend with 
    \Cref{fig2}}
    \label{fig3}
\end{figure}

As \Cref{fig2} shows, our proposed method outperforms all benchmarks compared against. 
Furthermore, the difference in performance between our proposed method and both 
round-robin \acro{BQ} and bridge sampling is significant at the 1\% significance level 
across all dimensions according to a one-sided paired $t$-test; the difference between 
our proposed method and reversible jump \acro{MCMC} is significant at the 1\% significance 
level for $d=$1, 2, and 3.

\subsection{Real-World Experiments}
\label{real}
Our real-world application is a model selection problem from the field of astrophysics. 
Given spectrographic observations of quasar emissions, astrophysicts are interested in 
inferring the existence of damped Lyman-$\alpha$ absorbers (\acro{DLA}s) between the 
quasar and earth. \acro{DLA}s are large gaseous clouds containing neutral hydrogen at 
high densities. The distribution of \acro{DLA}s throughout the universe is of interest as 
it informs galaxy formation models. Their location and size can be inferred from quasar 
spectra as they cause distinctive dips in the observed flux at well-defined wavelengths. 
\citet{garnett17} developed a model that specifies the likelihood that a quasar emission 
spectrum contains arbitrarily many \acro{DLA}s. Their likelihood model for $n$ \acro{DLA}s 
is parameterized by $2n$ parameters, two for each putative \acro{DLA}: one that 
corresponds to its size and one that corresponds to its distance from earth. 
\citet{garnett17} also specified a data-driven prior over these parameters; computing the 
model evidence for any number of \acro{DLA}s requires integrating the likelihood against 
this prior, resulting in an intractable integral.

In our experiments, we consider two candidate models for each quasar emission spectrum: 
the first corresponding to a single \acro{DLA} and the second corresponding to two 
\acro{DLA}s. In this experimental setting, we use the data-driven prior of 
\citet{garnett17} as the proposal distribution for the two additional parameters when 
transitioning from the single \acro{DLA} model to the two \acro{DLA} model. We select 20 
spectra from phase III of the Sloan Digital Sky Survey (\acro{SDSS--III}) 
\citep{eisenstein11}. We allot a budget of 150 total likelihood evaluations and initialize 
each \acro{BQ} based method with 25 randomly sampled likelihood observations from both 
model parameter spaces. We repeat the experiment 5 times for each spectra, using a 
different initialization for each trial. 

The diffeomorphism associated with our implementation of reversible jump \acro{MCMC} is 
again the identity function and the corresponding Jacobian factor is 1. The intractable 
integrals associated with our proposed method are estimated using quasi-Monte Carlo 
\citep{caflisch98}. We evaluated all methods on the absolute error of their estimates of 
the log odds or Bayes factor: $B_f=\nicefrac{z_1}{z_2}$ \citep{jeffreys61, kass95}, 
another potential quantity of interest in model selection tasks. We consider the absolute 
error instead of the fractional error as the target quantity is a log value. We make use 
of \citet{bartolucci06}'s work to translate the output Markov chain into a log odds 
estimate.

As \Cref{fig3} shows, our proposed method outperforms all benchmarks compared against. 
The difference in performance between our proposed method and both Monte Carlo based 
benchmarks is significant at the 1\% significance level according to a one-sided paired 
$t$-test.

\subsection{Model Choice Experiments}
\label{mc}
In situations where one is performing model choice as opposed to model averaging, the 
quantity of interest is not the model posterior probabilities but rather the model with 
the highest posterior probability, a related but different object. We considered an 
alternative acquisition function that targets this quantity, which we briefly present 
here. Given two candidate models $\model_1$ and $\model_2$, the goal in model choice is 
to determine the value of the indicator random variable $[z_1 > z_2]$, where we have 
adopted the Iverson bracket notation. Therefore, instead of considering the mutual 
information between $\ell_i(\theta_i)$ and $z_1$, one could conceivably consider the 
mutual information between  $\ell_i(\theta_i)$ and $[z_1 > z_2]$ directly. Formally, this
quantity can be expressed as
\begin{multline}
    I\bigl([z_1>z_2];\ell_i(\theta_i)\bigr)
    =\\
    H([z_1 > z_2]) - H \bigl([z_1 > z_2] \given \ell_i(\theta_i)\bigr).
    \label{eq18}
\end{multline}
Much like our proposed method, this alternative acquisition function searches over both 
models' parameter spaces for the next evaluation location:
\begin{equation}
    \theta_i^* = \argmax_{\theta_i} I\bigl([z_1>z_2];\ell_i(\theta_i)\bigr),    
    \label{eq19}
\end{equation}
for $i \in \{1,2\}$. Because the quantity $H([z_1>z_2])$ does not involve either 
$\ell_1(\theta_1)$ or $\ell_2(\theta_2)$, it can be safely ignored when searching for 
this maximum. Unfortunately, the second term in the mutual information \eqref{eq18} is 
intractable, much like the integrals in \eqref{eq14} and \eqref{eq15}. However, writing
\begin{multline*}
    H\bigl([z_1 > z_2] \given \ell_i(\theta_i)\bigr)
    =\\
    \int
    H\bigl(\Pr\bigl([z_1 > z_2] \given \ell_i(\theta_i)\bigr)\Bigr)
    \,
    p\bigl(\ell_i(\theta_i)\bigr)
    \intd \ell_i(\theta_i),
\end{multline*}
we may recognize the expression as a one-dimensional integral that can also be estimated 
numerically. 

\begin{figure}[!h]
    \begin{tabular}{c}
        \includegraphics{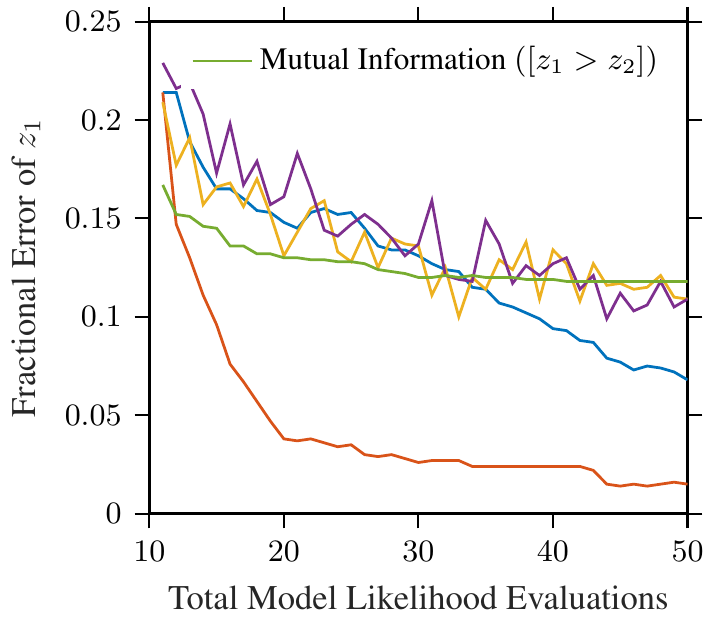} \\
        (a) $d=1$ \\[6pt]
        \includegraphics{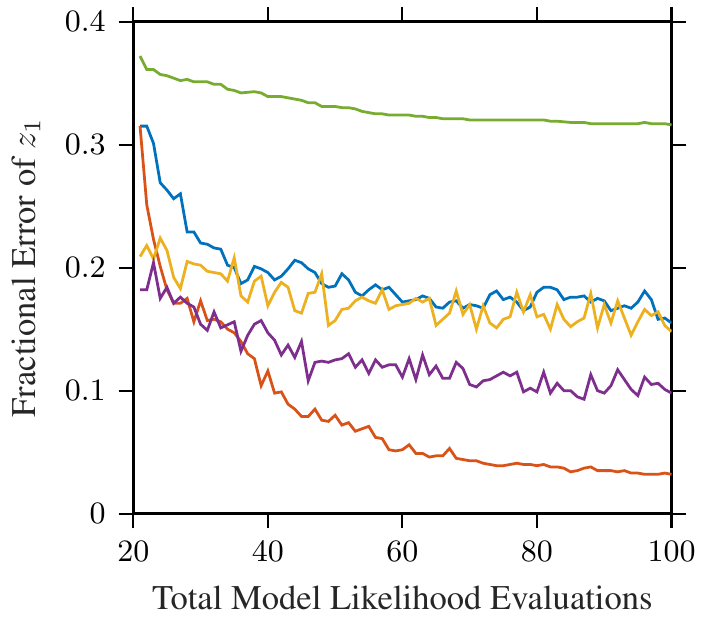} \\
        (b) $d=2$
    \end{tabular}
    \caption{Fractional errors of $z_1$ for all tested methods as well as an alternative 
    method considered specifically for the task of model choice. Please refer to 
    \Cref{fig2} for the omitted legend entries. }
    \label{fig4}
\end{figure}

\begin{figure}[!h]
    \includegraphics{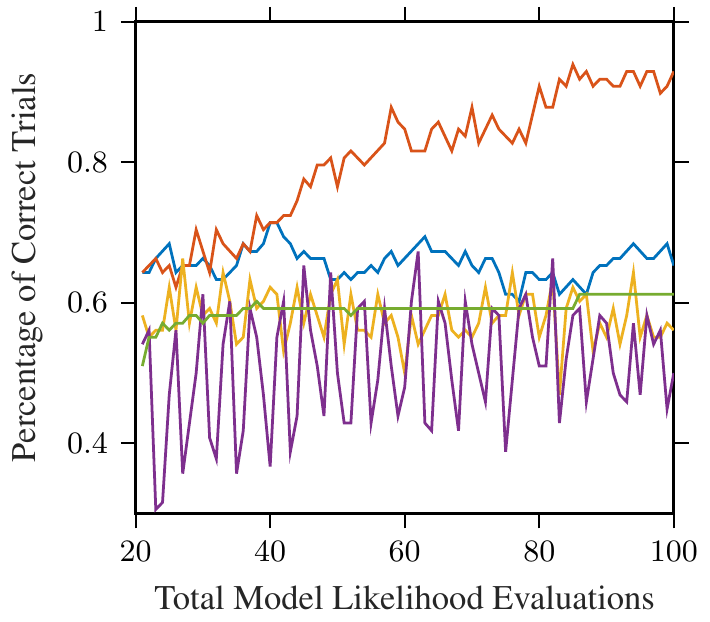}
    \caption{Fraction of trials where the ``correct" model or model with the higher ground 
    truth posterior probability would have been selected for all tested methods as well as 
    an alternative method considered specifically for the task of model choice. Please 
    refer to \Cref{fig2} for the omitted legend entries.}
    \label{fig5}
\end{figure}

Despite the arguably more rational choice to target $[z_1 > z_2]$ instead of $z_1$, this 
approach did not perform well in our experiments. \Cref{fig4} shows the performance of 
this method in the 1- and 2-dimensional synthetic experimental setting described above.
While this method is competitive with the benchmarks in 1-dimension, its performance is 
abysmal in the 2-dimensional setting; the performance continues to drops off as the 
number of dimensions increases. We also considered a different performance metric: the 
fraction of trials where the model with the higher ground truth posterior probability is 
correctly identified. It is reasonable to expect an acquisition function that targets 
$[z_1 > z_2]$ to outperform our method that targets $z_1$ when considering this metric. 
The results for the 2-dimensional synthetic experimental setting are shown in \Cref{fig5}.

We attribute the poor performance of this alternative acquisition function to the fact 
that the implied alternative objective of this acquisition function, the entropy of 
$[z_1 > z_2]$, is less stable than the entropy of $z_1$ and thus, this method has a 
tendency to become overly confident too quickly. If at any point one of the models 
achieves a much higher posterior model probability, then this method samples the model 
likelihoods at functionally uninformative locations until the budget of evaluations has 
been expended. This is because from the perspective of this alternative method, the 
objective has already been optimized: $\Pr(z_1 > z_2)$ will either be very close to zero 
or very close to one with very little uncertainty. In future work, we hope to improve the 
performance of this alternative method, potentially by targeting the random variable 
$z_1-z_2$ instead of $[z_1 > z_2]$ as we hypothesize that incorporating the magnitude of 
the difference will encourage continued exploration of the model parameter spaces. 
% !TEX root = ../main.tex

\section{Conclusion}
In this paper, we have presented a novel, \acro{BQ} based method for automated model 
selection. Our proposed method makes use of a novel acquisition function that targets the 
entropy of the posterior model probabilities, quantities specifically relevant to the task 
of model selection. This allows our method to actively sample locations across multiple 
model parameter spaces simultaneously. Our experiments conducted on real-world and 
synthetic datasets show that our proposed method can outperform both previously published 
\acro{BQ} approaches to model selection as well as Monte Carlo based model selection 
techniques in terms of achieving accurate posterior model probability estimates. 

An obvious extension of this work is to consider model selection tasks between $k>2$ 
models. Given $k$ candidate models $\{\model_1,\ldots,\model_k\}$, the mutual information 
between $l_i(\theta_i)$ and the vector of posterior model probabilities $[z_1,\ldots,z_k]$ 
can be computed using the same methodology in \ref{meth} i.e. by considering the joint 
Gaussian distribution between $l_i(\theta_i),a_1,\ldots,a_k,b_1,\ldots,b_k$ where 
$b_j=(z_j-1)a_j+z_j\sum_{j'\ne j}a_j'$. One exciting line of inquiry that we hope to study 
in future work concerns the design of trans-dimensional covariance functions. In certain 
settings, our assumption that the model evidences are independent does not accurately 
reflect our \emph{a priori} knowledge e.g. in model selection tasks with nested model 
parameters, we should expect model evidences to be at least slightly correlated. These 
types of relationships could be captured by a multi-task \acro{GP} \citep{cressie93, yu05} 
where the covariance between model likelihoods is learned alongside individual model 
likelihoods simultaneously. 

\newpage

\bibliography{main}
\bibliographystyle{icml2019}

\end{document}